\documentclass{article}

\usepackage[square]{natbib}

\usepackage[preprint]{nips_2018}

\usepackage[utf8]{inputenc} % allow utf-8 input
\usepackage[T1]{fontenc}    % use 8-bit T1 fonts
\usepackage{hyperref}       % hyperlinks
\usepackage{url}            % simple URL typesetting
\usepackage{booktabs}       % professional-quality tables
\usepackage{amsfonts}       % blackboard math symbols
\usepackage{nicefrac}       % closed symbols for 1/2, etc.
\usepackage{microtype}      % microtypography

\usepackage{times}
\usepackage{epsfig}
\usepackage{graphicx}
\usepackage{amsmath}
\usepackage{amssymb}

\usepackage{algorithm}
\usepackage{algorithmic}
\usepackage{amsthm}  
\usepackage{wasysym}
\usepackage[ruled,vlined,algo2e]{algorithm2e}

\usepackage{color}
\usepackage{dsfont}	
\usepackage{wrapfig}
\usepackage{footnote}
\usepackage{epstopdf}
\usepackage{enumitem}
\usepackage{subfigure}

\def\eg{\emph{e.g. }}
\def\ie{\emph{i.e. }}

\def\etc{\emph{etc. }} 
\def\vs{\emph{vs. }}
\def\wrt{\emph{w.r.t. }}
\def\etal{\emph{et al. }}

\makeatletter
\newcommand*{\rom}[1]{\expandafter\@slowromancap\romannumeral #1@}
\makeatother

\title{Deformable Part Networks}
\author{
	Ziming Zhang$^{\dag}$, Rongmei Lin$^{\ddag}$, Alan Sullivan$^{\dag}$ \\
	$^{\dag}$Mitsubishi Electric Research Laboratories (MERL), Cambridge, MA 02139-1955 \\
	$^{\ddag}$Department of Computer Science, Emory University, Atlanta, GA 30322\\
	\texttt{\{zzhang, sullivan\}@merl.com, rongmei.lin@emory.edu} \\
}

\begin{document}
	
\maketitle
	
\begin{abstract}

%In this paper we propose a novel network module {\em Deformable Capsule} (DeCap) that allows to efficiently detect multi-scale distinctive patterns on graphs using context-aware deformable models. We introduce graph-based local contexts into the capsules not only for detection explicitly, but also for grouping such patterns hierarchically and implicitly as fixed routing rules without learning. Inspired by the deformable part models for object detection in computer vision, we propose an algorithm with linear complexity by conducting detection {\em sequentially} from smaller contexts to larger ones. With these efforts our context-aware deep capsule networks (CADCN) can be naturally developed for graph data as input and trained efficiently. We demonstrate the empirical performance of our approach on classification tasks 

In this paper we propose novel Deformable Part Networks (DPNs) to learn {\em pose-invariant} representations for 2D object recognition. In contrast to the state-of-the-art pose-aware networks such as CapsNet \cite{sabour2017dynamic} and STN \cite{jaderberg2015spatial}, DPNs can be naturally {\em interpreted} as an efficient solver for a challenging detection problem, namely Localized Deformable Part Models (LDPMs) where localization is introduced to DPMs as another latent variable for searching for the best poses of objects over all pixels and (predefined) scales. In particular we construct DPNs as sequences of such LDPM units to model the semantic and spatial relations among the deformable parts as hierarchical composition and spatial parsing trees. Empirically our 17-layer DPN can outperform both CapsNets and STNs significantly on affNIST \cite{sabour2017dynamic}, for instance, by 19.19\% and 12.75\%, respectively, with better generalization and better tolerance to affine transformations.

\end{abstract}
	
\section{Introduction}
Very recently Sabour \etal \cite{sabour2017dynamic} proposed a new network architecture called CapsNet and a dynamic routing training algorithm which connects the capsules \cite{hinton2011transforming}, a new type of neurons that output vectors rather than scalars in conventional neurons, in two adjacent layers and groups similar features in higher layers. Later on Hinton \etal \cite{hinton2018matrix} proposed another EM-based routing-by-agreement algorithm for training CapsNet. In contrast to conventional convolutional neural networks (CNNs) that totally ignore the spatial relations between the filters, the intuition behind CapsNet is to achieve ``viewpoint invariance'' in recognizing objects for better generalization which is inspired by inverse graphics \cite{inverse-graphics}. Technically, CapsNet not only predicts classes but also encodes extra information such as geometry of objects, leading to richer representation than CNNs. For instance, in \cite{hinton2018matrix} $4\times 4$ pose matrices are estimated to capture the spatial relations between the detected parts and a whole. Empirically unlike CNNs the performance of CapsNet on real and more complex data has not been verified yet, partially due to the high computation that prevents it from being applicable widely.

%Deformable models \cite{terzopoulos1988deformable} is a well-known technique in computer vision community which provides the ability of modeling the variability of visual objects in terms of shape, texture or even imaging conditions. The basic idea in learning such models is to push the deformed model to match the data as well as possible (\ie loss functions) conditioned on some prior knowledge (\ie regularization). 

In fact exploring such invariant representations for object recognition has a long history in the literature of neural science as well as computer vision. For instance, in \cite{isik2013dynamics} Isik \etal observed that object recognition in the human visual system is developed in stages with invariance to smaller transformations arising before invariance to larger transformations, which supports the design of feed-forward hierarchical models of invariant object recognition. In computer vision part-based representation (\eg \cite{fischler1973representation}) is one of the most popular invariant object representations. In general part-based models consider an object as a graph where each node represents an object part and each edge represents the (spatial) relation between the parts. Conceptually part-based representation is view-invariant in 3D and pose-invariant (\ie translation, rotation, and scale) in 2D. Although the complexity of part-based models in inference on general graphs could be very high \cite{crandall2005spatial}, for tree structures such as star graphs this complexity can be linear to the number of parts \cite{felzenszwalb2012distance}.

Particularly Deformable Part Models (DPMs) \cite{felzenszwalb2010object} have achieved big success in object detection where, in general, we are interested in locating objects using bounding boxes. Based on the pictorial models \cite{felzenszwalb2005pictorial}, \ie star graphs, DPMs decompose an object into a collection of smaller parts, then detect these parts as well as modeling the geometric relations between the parts and the potential object center that are taken as latent variables. It has been demonstrated in the literature that DPMs are more robust to pose variations in objects than conventional detection methods such as template matching (\eg 2D convolution). %The computational complexity of DPMs in inference can be linear to the number of parts using distance transform \cite{felzenszwalb2012distance}.% provided that Gaussian distributions are utilized for modeling deformation in locations.
Recently in \cite{girshick2015deformable} DPMs is reinterpreted based on the operators in CNNs. 

As discussed above, both pose matrix and part-based representation are capable of capturing spatial relations among the parts in objects, but part-based representation (\eg DPMs) seems more suitable to be incorporated with conventional deep models. {\em So can we design a deep network based on part-based representation to learn pose-invariant object features as well as being trained efficiently?}

{\bf Contributions:}
In this paper we propose novel {\em Deformable Part Networks} (DPNs) to efficiently learn pose-invariant representation by estimating object poses in inference.%, which not only can capture the poses of objects as CapsNets, but also can be learned as CNNs. 

As a theoretical grounding of DPNs, we first propose a new challenging optimization problem in Sec.~\ref{sec:LDPM}, namely {\em Localized Deformable Part Models} (LDPMs), to learn these deformable parts as well as searching for the {\em best} pose of an object within multiple localized windows. The intuition of introducing localization into DPMs is that the deformation penalties of the parts are essentially dependent on the sizes of windows, and so is pose estimation. Meanwhile, localization enlarges the pose-searching space, leading to better capability of DPMs in modeling.

%to mimic pose invariance in a {\em discrete} manner, in contrast to the pose matrices in CapsNets that can be taken as estimation in a continuous space. The complexity in inference for LDPMs is linear to the sum of the areas within different scales used for pose estimation, roughly speaking.

%By taking LDPMs as basic modules, we further propose our DPNs that inherit the physical interpretation of LDPMs, \ie searching for best deformable part detections hierarchically to support recognition (see Sec.~\ref{sec:DPNs}). The complexity in searching, however, could be potentially very high for DPNs. To solve this problem, we propose an efficient algorithm to compute the deformation masks for larger scales with linear complexity. For instance, for a $5\times5$ kernel (each kernel defining a different scale), we keep computing the deformation using a $3\times3$ kernel twice in a {\em sequential} order. We prove that under certain conditions our algorithm can exactly recover (otherwise, approximate statistically) the deformation mask of a larger kernel using smaller kernels for more efficiency. In this way each learned deformation mask captures the spatial relation between the part and the center within a local context, which is functionally equivalent to a post matrix. %Our $\pi$-Nets can effectively learn the parts as well as their multi-scale deformation masks, and simultaneously group the part detections from lower to higher layers in a hierarchy.

The huge parameter space in solving LDPMs, however, brings significant computational challenges as well. Therefore, we propose DPNs as an efficient solver for LDPMs in Sec. \ref{sec:DPNs}. To regularize the parameter space, we propose using deformable part composition and spatial parsing trees that allow us to perform brute-force pose estimation in inference over all pixels as well as predefined windows. We also propose a new network operator, namely {\em Deformable Maxout} (DM), to learn the deformation penalties as well as doing inference with the same complexity as 2D convolution. % To accelerate the inference further, we approximate a DM with a larger window by a sequence of DMs with a smaller window. 
With such help our DPNs can be trained efficiently using stochastic gradient descent (SGD).

%Different from CapsNets, the learning of deformation masks in DPNs provides us a natural routing algorithm to hierarchically group the parts that share certain higher level properties such as semantics. Since all the operations in DPNs are differentiable, we can simply utilize stochastic gradient descent (SGD) based CNN solvers to train DPNs. 

%In a word, our DPNs can effectively learn the parts as well as their multi-scale deformation masks hierarchically for recognition. 
To demonstrate the effectiveness of DPNs, we review some related work in Sec. \ref{sec:rw} and conduct comprehensive experiments on MNIST \cite{lecun1998mnist}, affNIST \cite{sabour2017dynamic} and CIFAR-100 \cite{krizhevsky2009learning} in Sec. \ref{sec:exp}. We visualize the learned deformable part composition, spatial parsing trees, and feature distributions. Compared with some state-of-the-art networks, \ie VGG16 \cite{simonyan2014very}, ResNet32 \cite{he2016deep}, Spatial Transformer Networks (STNs) \cite{jaderberg2015spatial}, Deformable Convolutional Networks (DCNs) \cite{dai2017deformable} and  CapsNets, our DPNs can achieve better accuracy with better generalization to the number of training samples and better tolerance to affine transformations.

%As a result $\pi$-Nets learn to locate meaningful regions, represented by deformable parts in a hierarchy, that can support recognition. In Sec. \ref{sec:exp} we visualize this behavior as well as learned deformation masks empirically, along with comparison with CapsNets. 

To summarize, the main contributions of the paper are:
\begin{enumerate}[noitemsep, nosep]
\item[{\em C1.}] We propose novel Localized Deformable Part Models (LDPMs) that aims to learn the deformable parts as well as detecting the best object poses for recognition.
\item[{\em C2.}] We further propose novel Deformable Part Networks (DPNs) as an efficient solver for (regularized) LDPMs to learn pose-invariant object representations hierarchically. 
\item[{\em C3.}] We demonstrate the superiority of DPNs over the state-of-the-art networks.
\end{enumerate}

\section{Localized Deformable Part Models}\label{sec:LDPM}
%\subsection{Motivation}
{\bf Deformable Part Models (DPMs) \cite{felzenszwalb2010object}:} Given a window $x$ which is associated with the root filter, a DPM tries to learn a spatial configuration $z$ between the window center and part filters as well as a linear classifier $\mathbf{w}$ so that the following latent support vector machine (SVM) problem is optimization: 
\begin{align}\label{eqn:dpm}
\min_{\mathbf{w}\in\mathcal{W}}\left\{\sum_{(x,y)\in\mathcal{X}\times\mathcal{Y}}\ell\left(y, \max_{z\in\mathcal{Z}}\mathbf{w}^T\phi(x,z)\right)\right\},
\end{align}
where {\em w.l.o.g.} $y$ denotes a binary class label for $x$, $\mathcal{W}, \mathcal{X}, \mathcal{Y}, \mathcal{Z}$ denote the corresponding feasible spaces\footnote{For simplicity of the expressions, without explicit mentioning in this paper we assume that such feasible spaces satisfy the constraints on the variables such as regularization.}, $\ell$ denotes a loss function such as hinge loss, $\phi(x,z)$ denotes a structural feature vector involving appearance features of the (root and part) filters and deformation penalties of the parts (\eg the distance between the detected and learned locations of a part \wrt the root filter), and $(\cdot)^T$ denotes the matrix transpose operator. Note that DPMs can be applied to image recognition as well by taking each image as a window.

{\bf Pose Awareness:} Ideally a part-based representation is pose-invariant. Empirically, however, this nice property for recognition seems impossible to be achieved due to the visual ambiguity of parts. With the help of deformable parts, DPMs are more robust to such variance in appearance by taking spatial structures of the parts into account, \ie estimating poses of objects.

The ability of pose estimation in DPMs, however, is limited. First of all, the predefined and fixed scales of the root filters on image pyramid imply that DPMs can work well in the scenarios where at the coarse level the object poses need to be close to those defined in the root filters. Secondly the part deformations have no dependency on the object scales, making the pose estimation sensitive to the filter responses outside the object. 

\begin{wrapfigure}{r}{.5\linewidth}
	\vspace{-15pt}
	\begin{center}
		\includegraphics[width=\linewidth]{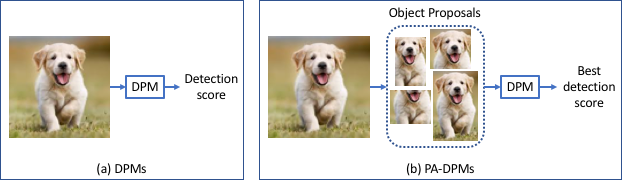}
		\vspace{-5mm}
		\caption{\footnotesize Illustration of comparison between DPMs and our LDPMs, \ie without \vs with localization.}
		\label{fig:padpm}
	\end{center}
	\vspace{-10pt}
	%  \vspace{1pt}
\end{wrapfigure}
In fact, pose estimation heavily depends on {\em object scales}. Imagining two same faces but with different scales, the part detectors for eyes, nose, mouth, \etc as well as their deformations in the faces may be represented differently. For instance, the eye detector for the larger face may be visually bigger (\ie covering more pixels) than that for the smaller face in order to detect the same ``eye'' semantically. Therefore, to improve pose estimation in DPMs, we propose using {\em localization} to search for the best object scale in a window, \ie object proposals (\eg \cite{zhang2016object}), a widely used technique in object detection such as RCNN \cite{girshick2014rich}. As illustrated in Fig. \ref{fig:padpm}, rather than feeding the window of dog into a DPM, we extract proposals from the window and then feed them into a DPM for recognition. Then our LDPMs can capture the best pose with highest detection score among all the proposals.

{\bf Formulation:}
Similar to Eq. \ref{eqn:dpm}, we propose another latent SVM problem for LDPMs as follows:
\begin{wrapfigure}{r}{.25\linewidth}
	\vspace{-5pt}
	\begin{center}
		\includegraphics[width=\linewidth]{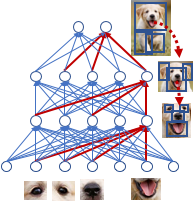}
		\vspace{-5mm}
		\caption{\footnotesize Illustration of deformable part composition.}
		\label{fig:tree}
	\end{center}
	\vspace{-10pt}
	%  \vspace{1pt}
\end{wrapfigure}
\begin{align}\label{eqn:padpm}
\min_{\mathbf{w}\in\mathcal{W}}\left\{\sum_{(x,y)\in\mathcal{X}\times\mathcal{Y}}\ell\left(y, \max_{(h,z)\in\mathcal{H}\times\mathcal{Z}}\Big\{\mathbf{w}^T\phi(x,h,z)\Big\}\right)\right\},
\end{align}
where $h\in\mathcal{H}$ denotes a proposal within a window $x$, and $\phi(x,h,z)$ denotes the structural feature vector conditioned on latent variables $h$ and $z$. % and function $b:\mathcal{H}\rightarrow\mathbb{R}\in\mathcal{B}$ outputs a bias term as calibration based on the input patch size that accounts for the change in scale. For instance, if two patches $h_1, h_2$ share the same size, then $b(h_1)=b(h_2)$.
Note that variable $h$ accounts for the {\em multi-scale} localization. %Due to the characteristics of digital images, our LDPMs essentially discretize a pose by aligning it with pixels and encoding it in latent variables $(h, z)$.

In contrast to DPMs where the inference is conducted in the 2D image space, our LDPMs perform the inference in a 4D space consisting of different proposals as its points. The higher dimensionality brings not only new challenges into learning, but also the flexibility in modeling that may lead us to some solutions with better accuracy and convergence.

% {\bf Discussion:}
% Due to the spatial quantization in images, our LDPMs essentially discretize the pose space based on pixels, and encode the parameters in $\mathbf{w}$. The usage of $b$ is for distinguishing the significance of detections at different scales. We can also learn a model (\ie $\mathbf{w}$) for each scale, namely multi-scale learning model, but this will lead to much higher model complexity. The computational complexity of both models in inference is the same. 
% Empirically we find that the multi-scale learning model can achieve slightly better performance, but the difference will be marginalized with the increase of the network depth.

\section{Deformable Part Networks}\label{sec:DPNs}
\subsection{Mathematical Modeling}
\textbf{Deformable Part Composition \& Spatial Parsing Tree as Regularization:}
In general the search space for pose estimation can be as huge as $O(N^P)$ where $N$ is the image size and $P$ is the number of parts. In order to explore such huge space efficiently, we propose using tree-structural composition to model the semantic dependency between the parts explicitly. As shown in \cite{mhaskar2016deep} deep models can approximate complex functions (\eg the decision function $\mathbf{w}^T\phi(x,h,z)$ in Eq.~\ref{eqn:padpm}) more efficiently and compactly than shallow models. Fig. \ref{fig:tree} illustrates an example of our deformable part composition, where each node represents a detector from object classes down to the parts and the red edges denote the dependency between the fired detectors. To facilitate the learning we share all the parts among the classes as suggested in \cite{torralba2004sharing} that may lead to better model complexity as well. The responses of fired detectors at a lower layer are passed to the detectors at a higher layer, together with their supporting regions in the image that form spatial parsing trees for estimating poses. These parsing trees actually follow the methodology in DPMs to control the appearance variances of parts based on the semantic hierarchy. Note that each detector can have multiple fires with different windows in image domain.

We would like to learn a model that can learn the deformable part composition as well as inferring the spatial parsing tree in each image as pose estimation to generate pose-invariant representations.

{\bf Key Notations:}
We denote $\{(x,y)\}\subseteq\mathcal{X}\times\mathcal{Y}$ as the training data with image $x$ and its label $y$, $i\in x$ as the $i$-th pixel in $x$, and $h(i)\in\mathcal{H}$ as a window centered at $i$ that consists of a collection of pixels. We also predefine an $N$-layer deformable part composition where there exist $d_n (n\in[N])$ nodes in the $n$-th layer. Further we denote $\psi(x,h(i),n)\in\mathbb{R}^{d_n}$ as the scoring vector at pixel $i$ within window $h(i)$ in image $x$ using the detectors in the $n$-th layer, $\psi(x,i,n)\in\mathbb{R}^{d_n}$ accordingly for window $h(i)=\{i\}$, and $\alpha(h,j,n), \beta(h,j,n)\in\mathbb{R}^{d_n}$ as the deformation penalty parameters at pixel $j\in h$ within window $h$ for the detectors in the $n$-th layer. We also denote matrix $\mathbf{W}_{n}\in\mathbb{R}^{d_{n-1}\times d_n}, \forall n$ as the semantic composition weights between the $(n-1)$-th and $n$-th layers, $\ell$ as the loss function for recognition, $\sigma$ as the activation function such as ReLU \cite{nair2010rectified} for firing detectors, and all the $\max$ operators in the following sections are entry-wise. 

{\bf Formulation for Brute-Force Pose Estimation:}
Based on deformable part composition and spatial parsing trees, we manage to convert the pose estimation problem at inference time in Eq.~\ref{eqn:padpm} to localizing a proper window per pixel where an LDPM is conducted for pose estimation.

To do so, we first decompose the feature vector $\phi(x,h(i),z)=\{\psi(x,h(i),n)\}_{1\leq n\leq N}$ as a collection of scoring vectors using the detectors per layer. Then we explicitly define $\psi(x,h(i),n)$ as follows:
\begin{align}\label{eqn:psi}
\psi(x,h(i),n) = \max_{j\in h(i)}\Big\{\alpha(h,j,n)\otimes\psi(x,j,n) \oplus \beta(h, j, n)\Big\}, \forall x, \forall i, \forall n,
\end{align}
where $\otimes, \oplus$ denote the entry-wise product and summation between two vectors, respectively. %Note that Eq. \ref{eqn:psi} is essentially equivalent to the inference in DPMs. 
Different from previous works such as \cite{felzenszwalb2010object, ouyang2015deepid} where deformation penalties are parameterized based on explicit deformation features such as distances, we directly model these penalties using latent learnable functions $\alpha, \beta$ that take window size, pixel location, and detectors as input and output vectors to penalize the deformation \wrt the window center.

Now based on Eq. \ref{eqn:padpm} and Eq. \ref{eqn:psi} we propose a new specific LDPM formulation as follows\footnote{For simplicity in formulation we keep the image sizes unchanged during both learning and inference.}:
% \begin{align}\label{eqn:padnet}
% \min_{\substack{\{\mathbf{W}_{n,h}\}\subseteq\mathcal{W} \\ g\in\mathcal{G}}}\sum_{(x,y)\in\mathcal{X}\times\mathcal{Y}}\ell\Big(y, \max_i\psi(x,i,N)\Big),
% \mbox{s.t.} \, \psi(x,i,n) = \sigma\left(\max_{h(i)\in\mathcal{H}}\mathbf{W}_{n,h}^T\psi(x,h(i),n-1)\right), \forall n,
% \end{align}
\begin{align}\label{eqn:padnet}
& \min_{\{\mathbf{W}_{n}\}\subseteq\mathcal{W}, \alpha\in\mathcal{A}, \beta\in\mathcal{B}}\sum_{(x,y)\in\mathcal{X}\times\mathcal{Y}}\ell\Big(y, \max_i\psi(x,i,N)\Big), \\
& \mbox{s.t.} \; \psi(x,i,n) = \sigma\left(\max_{h(i)\in\mathcal{H}}\mathbf{W}_{n}^T\psi(x,h(i),n-1)\right), \forall x, \forall i, \forall n, \nonumber
\end{align}
where $\psi(x,i,0)\in\mathbb{R}^{d_0}, \forall x, \forall i$ denotes the raw image feature vector at pixel $i$ in image $x$. As discussed in Sec.~\ref{sec:LDPM}, the detectors for the same semantic concept (\eg eye) may be dependent on window sizes. To account for the semantic consistency within different windows, here we intentionally employ multiple instance learning (MIL), parameterized by $\mathbf{W}_n, \forall n$, to select the best window.

By simultaneously considering all the pixels and window sizes defined in $\mathcal{H}$ for solving the specific LDPM problem in Eq. \ref{eqn:padnet}, we indeed search for the best poses of objects in a brute-force manner at inference time in a 4D window space. Together with variables $\mathbf{W}_{n}, \forall n$ and latent functions $\alpha, \beta$ that conduct the part detection and deformation in multi-scale scenarios, our approach can manage to estimate the best poses and thus generate pose-invariant representations for object recognition.

\subsection{DPNs: an Efficient LDPM Solver}
\begin{wrapfigure}{r}{.59\linewidth}
	\vspace{-15pt}
	\begin{center}
		\includegraphics[width=\linewidth]{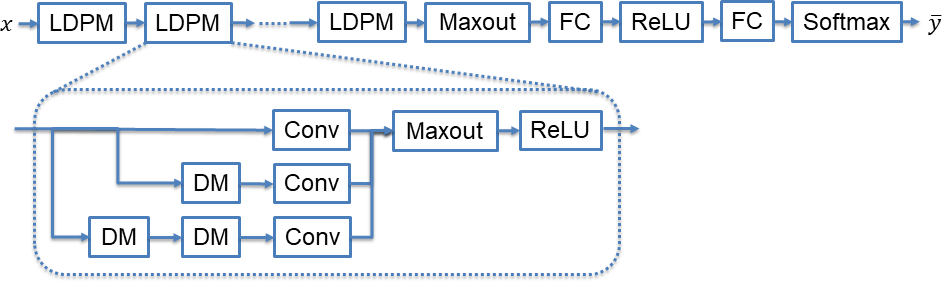}
		\vspace{-5mm}
		\caption{\footnotesize Illustration of DPN architecture.}
		\label{fig:padnet}
	\end{center}
	\vspace{-10pt}
	%  \vspace{1pt}
\end{wrapfigure}
The outputs of $\alpha, \beta$ in Eq. \ref{eqn:psi} can be considered as two sets of filters, similar to those in 2D convolution. To do the inference in Eq. \ref{eqn:psi}, we propose a new network operator, namely {\em Deformable Maxout} (DM), by applying $\otimes, \oplus$ sequentially over different channels, which has the same computational complexity as 2D convolution.

Considering the computational efficiency, we refer to the inception in GoogLeNet \cite{szegedygoing} and predefine the feasible window set $\mathcal{H}$ for measuring part deformation as a collection of $1\times1$, $3\times3$ and $5\times5$. With the increase of network depth, larger windows (\eg $7\times 7$) will be covered as the receptive fields of neurons, and due to the characteristic of deformation any arbitrary window within the receptive field can be potentially localized as a part. This implicitly favors the brute-force pose estimation.

{\bf Architecture:} 
We illustrate our DPNs in Fig.~\ref{fig:padnet} as an efficient solver for LDPMs defined in Eq.~\ref{eqn:padnet}. In feedforward procedure, each LDPM unit computes the detection scores as well as conducting pose estimation where the filters in convolution layers are {\em shared}, and the maxout layer \cite{goodfellow2013maxout} is over $\mathcal{H}$. The maxout layer outsize LDPM units is over the image domain. %The  operator is used to introduce nonlinearity that indicates whether a detector, defined by the kernel weights in the 2D convolution, is fired or not. 
The convolution layers and DM layers are responsible for learning $\mathbf{W}_n, \forall n$ and $\alpha, \beta$ in Eq. \ref{eqn:padnet}, respectively, that are updated in back-propagation. %Within each LDPM unit even though the window set for consideration is predefined as in Eq. \ref{eqn:H}, the optimal window for inference can be localized in an arbitrary size that is no larger than $(2M+1)\times(2M+1)$, duo to the characteristic of DPMs. Meanwhile, the deeper a LDPM unit is in the network, the larger the receptive field of each detector is, which corresponds to our spatial parsing tree.
Since the DM operator is differentiable with the same complexity as 2D convolution, our DPNs can be trained as efficiently as CNNs.

{\bf Implementation:} 
Without fine-tuning we set both filter size in all the convolution layers and window size in DM layers to $3\times3$. Similar to \cite{szegedy2016rethinking}, we use two sequential DM operations to approximate the output of DM with $5\times5$ windows. %We take the valid matrix of each DM layer as output and pad all the matrices with zeros as the inputs for the max layer. 
We initialize the network based on \cite{he2015delving}, and optimize it using ADAM \cite{kingma2014adam}. Batch normalization \cite{ioffe2015batch} can be employed as well. Particularly as demonstration, in our experiments we set the numbers of kernels in LDPM units as well as the fully connected (FC) layer to $32\rightarrow64\rightarrow128\rightarrow256\rightarrow512\rightarrow1024\rightarrow10$, respectively. This is equivalent to a 17-layer CNN by taking the longest path in the network from input to output and counting DM, Conv, and FC only along the path. Considering the datasets, we apply image downsampling twice by 2 after the LDPM units with 64 and 256 filters, respectively, for better computational efficiency with little impact on pose estimation as the DM layers will compensate for the inaccuracy in localization.

Empirically we find that both the depth of the networks and the width of LDPM units have impact on the performance. In general the deeper and wider the network, the better the accuracy with slower running speed. For instance, on affNIST with the increase of the width from 3 (as shown currently in Fig. \ref{fig:padnet}) to 7 that capture larger windows (\ie from $7\times7$ to $13\times 13$, step by 2), the accuracy is improved by about 2\%. If the width is reduced to 1 (\ie only the first branch in LDPM units left), the accuracy drops by about 8\%. Relatively the depth of the networks are more important than the width of LDPM units in order to achieve good performance. This is reasonable as larger windows can be captured by the neurons at the higher layers in the networks.

\section{Related Work}\label{sec:rw}
Besides DPMs and CapsNets, we summarize some other related work as follows.

{\bf Hierarchical Deformable Part Models (HDPMs):}
Felzenszwalb \etal \cite{felzenszwalb2010cascade} proposed a cascade detection algorithm based on partial hypothesis pruning for accelerating DPMs that can be defined by a grammar formalism. Zhu \etal \cite{zhu2010latent} proposed a mixture of three-layer latent hierarchical tree models for object detection and learned it using an incremental concave-convex procedure (iCCCP). Ghiasi and Fowlkes \cite{ghiasi2014occlusion} proposed a two-layer HDPM for modeling occlusion in faces that achieved state-of-the-art performance on benchmarks for occluded face localization. Tian \etal \cite{tian2012exploring} proposed a three-layer hierarchical spatial model that can capture an exponential number of poses with a compact mixture representation on each part using exact inference. Wu \etal \cite{wu2016learning} proposed a And-Or car detection model to embed a grammar for representing large structural and appearance variations in a reconfigurable hierarchy that is trained using Weak-Label Structural SVM. 

In contrast to these works, our LDPMs can involve much deeper deformable part hierarchy and learn these parts automatically rather than manual design based on certain prior knowledge. Further we propose using DPNs as our efficient solver for LDPMs that can naturally learn the semantic and spatial relations among the parts in the hierarchy. 

{\bf Pose-Aware Networks:}
%Wan \etal \cite{wan2015end} proposed a network by integrating the convolution, DPM, and non-maximal suppression (NMS) operations where DPMs are used to capture different views.
In terms of applications, poses are usually considered and encoded into networks for the recognition of specific object classes such as faces \cite{masi2016pose} or human \cite{humanpose2015sp, wei2016convolutional, kumar2017pose}. For instance, Wei \etal \cite{wei2016convolutional} proposed Convolutional Pose Machines (CPMs) that provide a sequential prediction framework for learning rich implicit spatial models for human pose estimation. The body parts are well defined visually with clear spatial supports but without semantically composite relations. Differently our DPNs are developed for estimating the poses of general objects based on deformable part composition and spatial parsing trees.

In terms of functionality, some pose-aware network operations or modules as plug-in are proposed for existing networks. For instance, dilated convolution \cite{yu2015multi} supports exponential expansion of the receptive field (\ie window) without loss of resolution or coverage and thus can help networks capture multi-scale information. Deformable Convolutional Networks (DCNs) \cite{dai2017deformable} proposed a more flexible convolutional operator that introduces pixel-level deformation, estimated by another network, into 2D convolution. Spatial Transformer Networks (STNs) \cite{jaderberg2015spatial} learn pose-invariant representations by sequential applications of a localization network, a parameterized grid generator and a sampler.

In contrast, we propose a new DM operator that can be used to learn deformation penalties for parts as well as conducting the inference for DPMs within fixed windows. Based on DM we propose LDPM units as network modules in DPNs to solve the optimization problem in LDPMs by estimating object poses as well as predicting class labels, leading to pose-invariant representations for recognition.

{\bf DPN \vs GoogLeNet \& ResNet \cite{he2016deep}:}
In particular, in terms of architecture our LDPM units in DPN are related to the inception module in GoogLeNet and ResNet. Both LDPM and inception are able to capture multi-scale information. Differently the receptive fields in inception are fixed by the sizes of convolutional filters, while LDPM manages to locate arbitrary windows with best part detection scores within each receptive field defined by DM. Compared with ResNet, LDPM can have skip connections as well by removing the convolutional layers. Differently LDPM takes entry-wise maximum over different receptive fields rather than summation, leading to feature selection in LDPM.

\section{Experiments}\label{sec:exp}
We test and compare our DPN with some state-of-the-art networks with similar model complexity to ours, \ie VGG16\footnote{https://github.com/geifmany/cifar-vgg}, ResNet32\footnote{https://github.com/tensorflow/models/tree/master/research/resnet}, STN\footnote{https://github.com/kevinzakka/spatial-transformer-network}, DCN\footnote{https://github.com/felixlaumon/deform-conv}, and CapsNet\footnote{https://github.com/naturomics/CapsNet-Tensorflow}. We use three benchmarks, namely MNIST, affNIST, and CIFAR-100. Particularly affNIST is created for testing the tolerance of an algorithm to affine transformation (\ie translation, rotation, shearing and scaling). On MNIST we follow standard procedure to train the networks using the 60K training/validation samples and test them using the 10K test samples. On affNIST we follow \cite{sabour2017dynamic} to create a new training set of 60K samples using original MNIST training/validation samples with {\em random translation only}, and train all the networks using it, then test them using the 10K test samples in affNIST which involve {\em random affine transformations}. To facilitate the training, we resize the images to $28\times28$, same as MNIST. On CIFAR-100 we utilize the pre-processing code\footnote{https://github.com/tensorflow/models/blob/master/tutorials/image/cifar10} for network training using the 50K training samples, and test the networks using the 10K test samples.

\setlength{\tabcolsep}{2pt}
\begin{wraptable}{r}{9.3cm}\small
	\vspace{-15pt}
	\begin{center}    
		\begin{tabular}{|c||c|c|c|c|c|c|}
			\hline & {\bf Ours (DPN17)} & VGG16 & ResNet32 & STN & DCN & CapsNet \\ 
            \hline MNIST & 99.48 & {\bf 99.71} & 99.23 & 99.26 & 99.45 & 99.66 \\
            \hline affNIST & {\bf 97.26} & 93.12 & 92.89 & 84.51 & 90.32 & 78.07 \\
            \hline CIFAR-100 & {\bf 70.96} & 70.48 & 68.10 & 66.18 & 67.39 & - \\
			\hline
		\end{tabular}
	\end{center}
    \vspace{-2mm}
	\caption{\footnotesize Best accuracy (\%) comparison on different datasets, where ``-'' indicates that we cannot achieve reasonable performance.}\label{tab:acc}
    \vspace{-3mm}
\end{wraptable}
\textbf{Better Performance:} We summarize the accuracy comparison in Table \ref{tab:acc}. As we see on all the datasets our DPN significantly outperforms the three pose-aware networks, \ie STN, DCN, and CapsNet. Compared with well designed CNN based networks, \ie VGG16 and ResNet32, DPN is always comparable. We believe that such observations are mainly the outcomes of learning deformation in DPN that helps capture the part configurations of objects efficiently rather than ``memorizing'' the training instances. 

\begin{wrapfigure}{r}{.4\linewidth}
	\vspace{-15pt}
	\begin{center}
		\includegraphics[width=\linewidth]{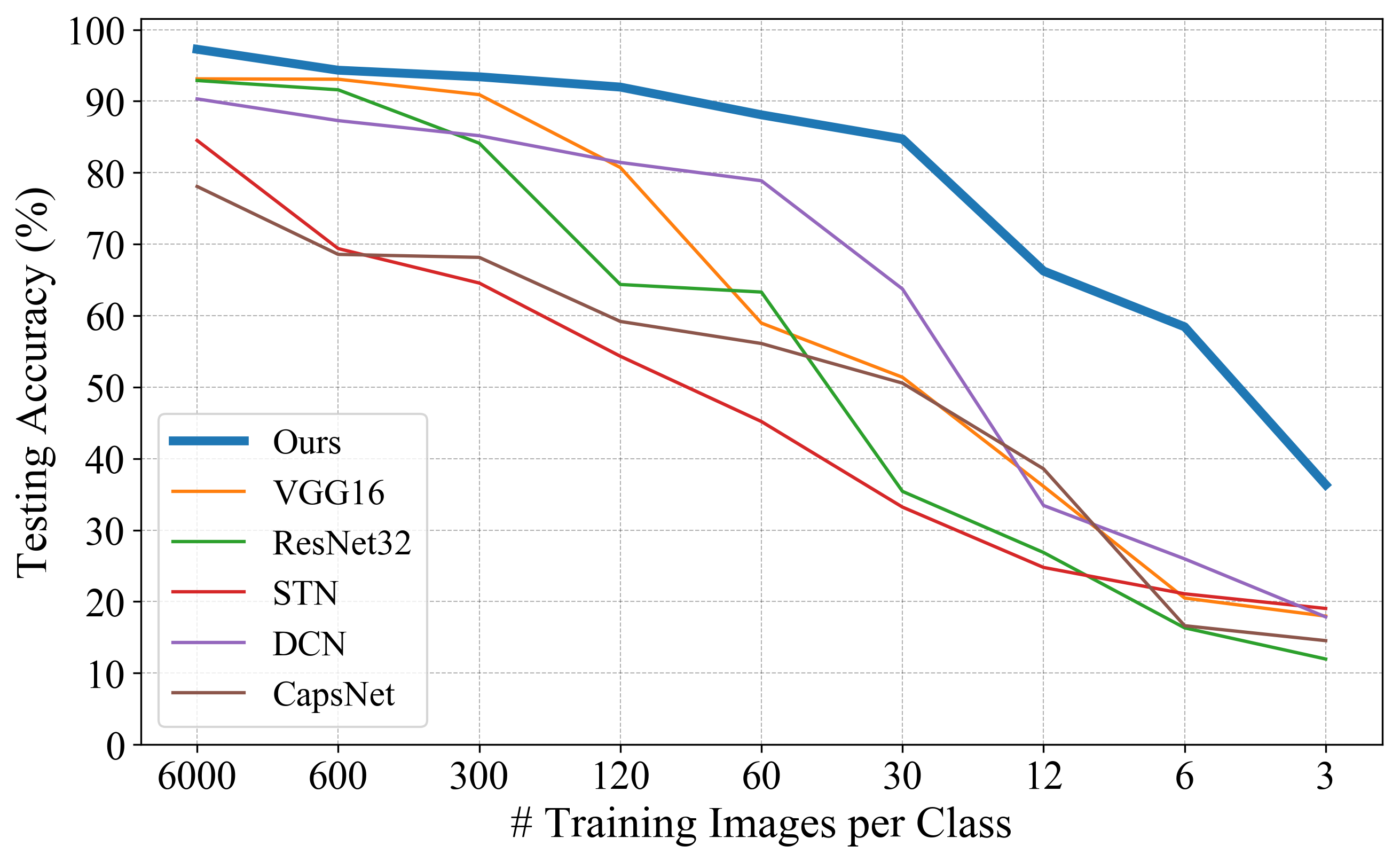}
		\vspace{-7mm}
		\caption{\footnotesize Performance analysis on affNIST.}\label{fig:test}
        \vspace{-5mm}
	\end{center}
\end{wrapfigure}
\textbf{Better Generalization:} To verify our hypothesis, we conduct a performance analysis over the number of training images per class on affNIST as shown in Fig.~\ref{fig:test}. Overall the accuracy of all the methods becomes worse when the number decreases. DPN, however, behaves much more robustly. In the extreme case where there are only 3 images per class for training, DPN can achieve {\bf 36.35\%} that is {\bf 17.31\%} improvement over the second best. Considering our model complexity that is much higher than ResNet32 and DCN, in such extreme cases DPN should perform worse due to the higher risk of overfitting. Surprisingly, however, this is not true on affNIST. Since the key difference between DPN and other networks is that we utilize DPMs to estimate poses, we then conclude that the learned deformable part configurations do help us recognize new digits that are never seen before, leading to better generalization.

\begin{figure}[t]
	\begin{center}
		\includegraphics[width=1\linewidth]{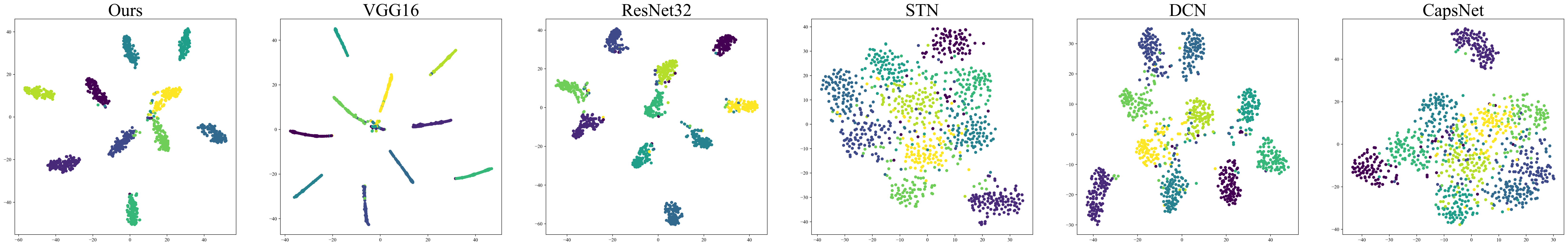}
		\vspace{-5mm}
		\caption{\footnotesize Feature distribution comparison on affNIST test data using t-SNE \cite{maaten2008visualizing}, one color per class.}\label{fig:distributions}
	\end{center}
    \vspace{-5mm}
\end{figure}
\begin{figure}[t]
	\begin{center}
		\includegraphics[width=1\linewidth]{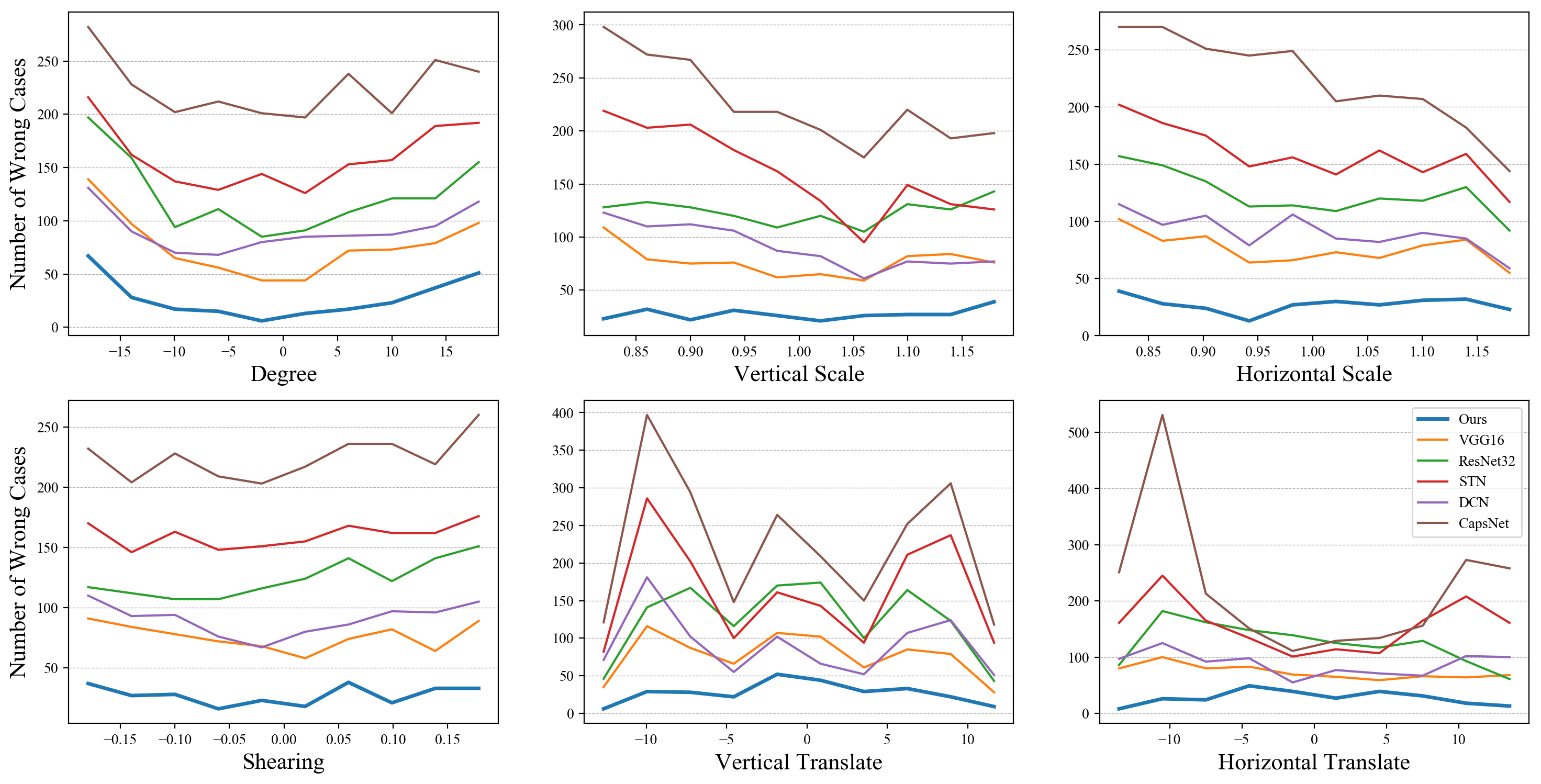}
		\vspace{-5mm}
		\caption{\footnotesize Analysis of failure cases on affNIST test data over different transformations.}\label{fig:analysis}
	\end{center}
    \vspace{-5mm}
\end{figure}

\setlength{\tabcolsep}{2pt}
\begin{wraptable}{r}{8.6cm}\small
	\vspace{-15pt}
	\begin{center}    
		\begin{tabular}{|c||c|c|c|c|c|c|}
			\hline & {\bf Ours} & VGG16 & ResNet32 & STN & DCN & CapsNet \\ 
            \hline Intra-cls. dis. & {\bf 4.79} & 6.20 & 6.57 & 7.31 & 6.19 & 8.20 \\
            \hline Inter-cls. dis. & {\bf 24.78} & 21.62 & 23.23 & 17.25 & 19.09 & 17.39\\
            \hline Intra/Inter & {\bf 0.19} & 0.28 & 0.28 & 0.42 & 0.32 & 0.47 \\
			\hline
		\end{tabular}
	\end{center}
    \vspace{-2mm}
	\caption{\footnotesize Comparison on intra-class and inter-class distances in Fig.~\ref{fig:distributions}.}\label{tab:dis}
    \vspace{-3mm}
\end{wraptable}
\textbf{Better Tolerance to Affine Transformation:} To illustrate the differences of learned features among the networks, we show the feature distribution comparison in Fig. \ref{fig:distributions} using t-SNE. For each network we extract the features that are fed into the classifier module directly. For instance, for DPN we extract the 1024D features. As we see the our DPN features can form more compact and separable clusters than the other pose-aware networks. To quantify the distributions, we measure the intra-class (\wrt compactness) and inter-class (\wrt separability) distances in the 2D space in Fig. \ref{fig:distributions} and list the comparison in Table~\ref{tab:dis}. Clearly our intra-class distance is much smaller than the others while inter-class distance is larger, leading to more discriminative (and probably pose-invariant) distributions for recognition. % These results also indicate that DPN features are more likely to being . %Due to page limit, here we do not show the distributions for VGG16 (6.2498) and ResNet32 (6.7378), but our distance measure is still significantly smaller. 

As shown in Fig. \ref{fig:analysis} we analyze the failure cases on affNIST as well. We quantize each parameter for affine transformation into 1 of 10 bins accordingly, and the total numbers of failure cases per network in the 6 subfigures are the same. As we see our DPN is the most robust to all of the transformations, leading to fewest failure cases among all the networks. Among different   transformations DPN is more sensitive to rotation (see the top-left subfigure), as more degree for rotation is, more failure cases occur. For the others the distributions appear more or left flat. This is mainly because rotation has larger impact on the learning of parts as well as their spatial configurations, making the pose estimation less reliable.

\begin{figure*}[t]
   \begin{minipage}{0.495\textwidth}
     \centering
     \includegraphics[width=.85\linewidth]{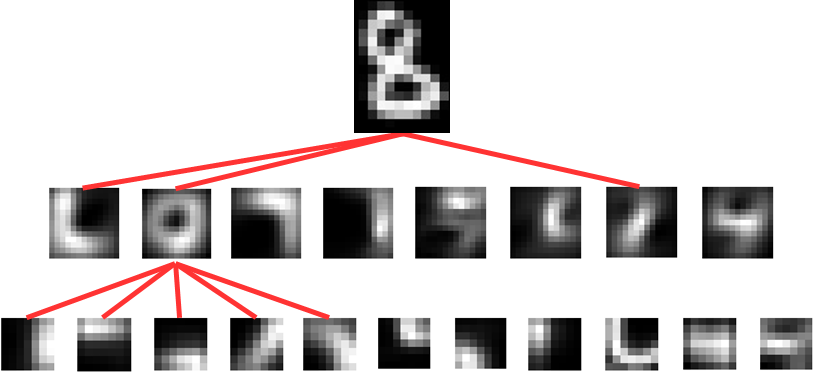}     
   \end{minipage}\hfill
   \begin{minipage}{0.495\textwidth}
     \centering
     \includegraphics[width=.85\linewidth]{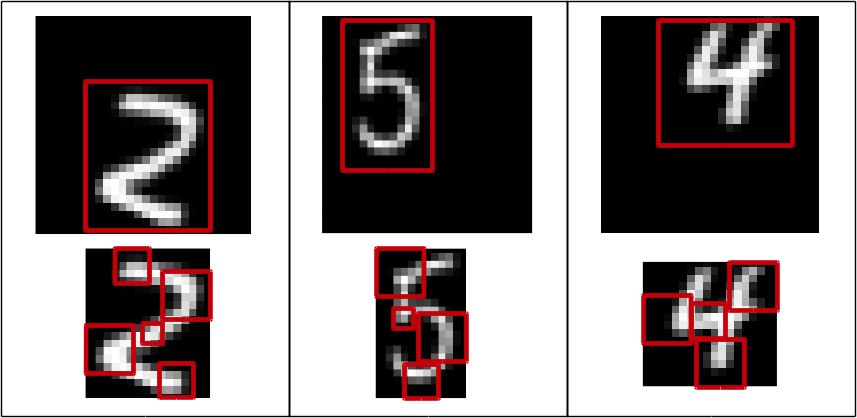}     
   \end{minipage}
   \vspace{-1mm}
   \caption{\footnotesize Deformable part visualization on affNIST ({\bf left}) semantically and ({\bf right}) spatially.}\label{fig:aff_train}
   \vspace{-1mm}
\end{figure*}

\textbf{Deformable Part Visualization:}
To better understand our DPN, we visualize some learned deformable parts in Fig. \ref{fig:aff_train}. On the left, as an example we show the hierarchical decomposition of a digit ``8'' using learned parts, where the red lines denote the edges with positive weights in the hierarchy. Here all the parts are rescaled to a same size per layer, and we can only show the top two layers because it becomes difficult to clearly visualize smaller parts in the lower layers with few pixels. On the right, we visualize the spatial configures of some parts in the top two layers of the spatial parsing trees for digit ``2'', ``5'', and ``4'' respectively. As we see the spatial configurations of deformable parts can indeed locate the digits nicely with different windows, within which the scales of the detected parts may vary significantly. These observations come from the characteristics of LDPM units that can search for the best object poses over all possible pixels and predefined windows.

\begin{figure*}[t]
   \begin{minipage}{0.495\textwidth}
     \centering
     \includegraphics[width=.85\linewidth]{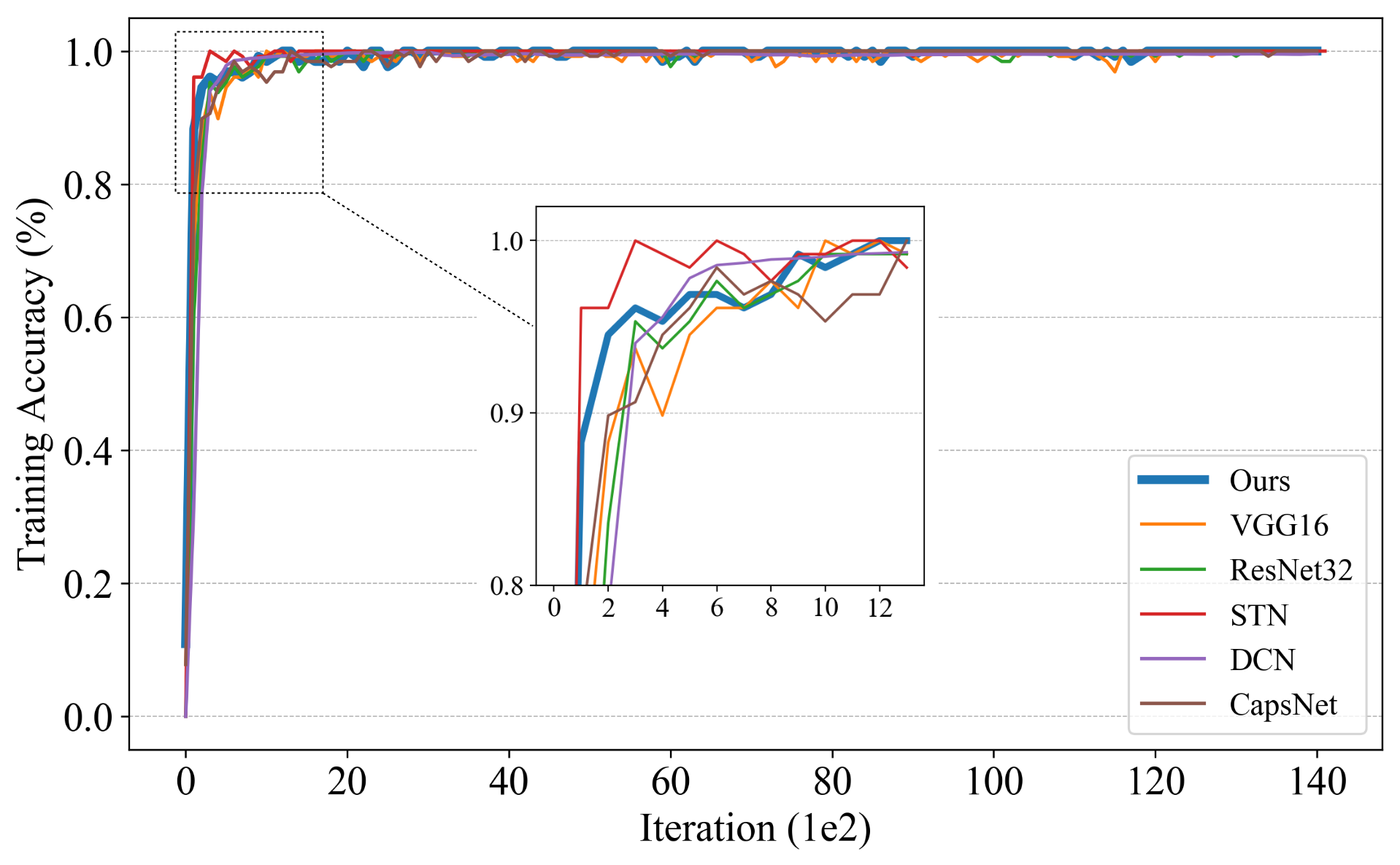}
     %\vspace{-10pt}     
   \end{minipage}\hfill
   \begin{minipage}{0.495\textwidth}
     \centering
     \includegraphics[width=.85\linewidth]{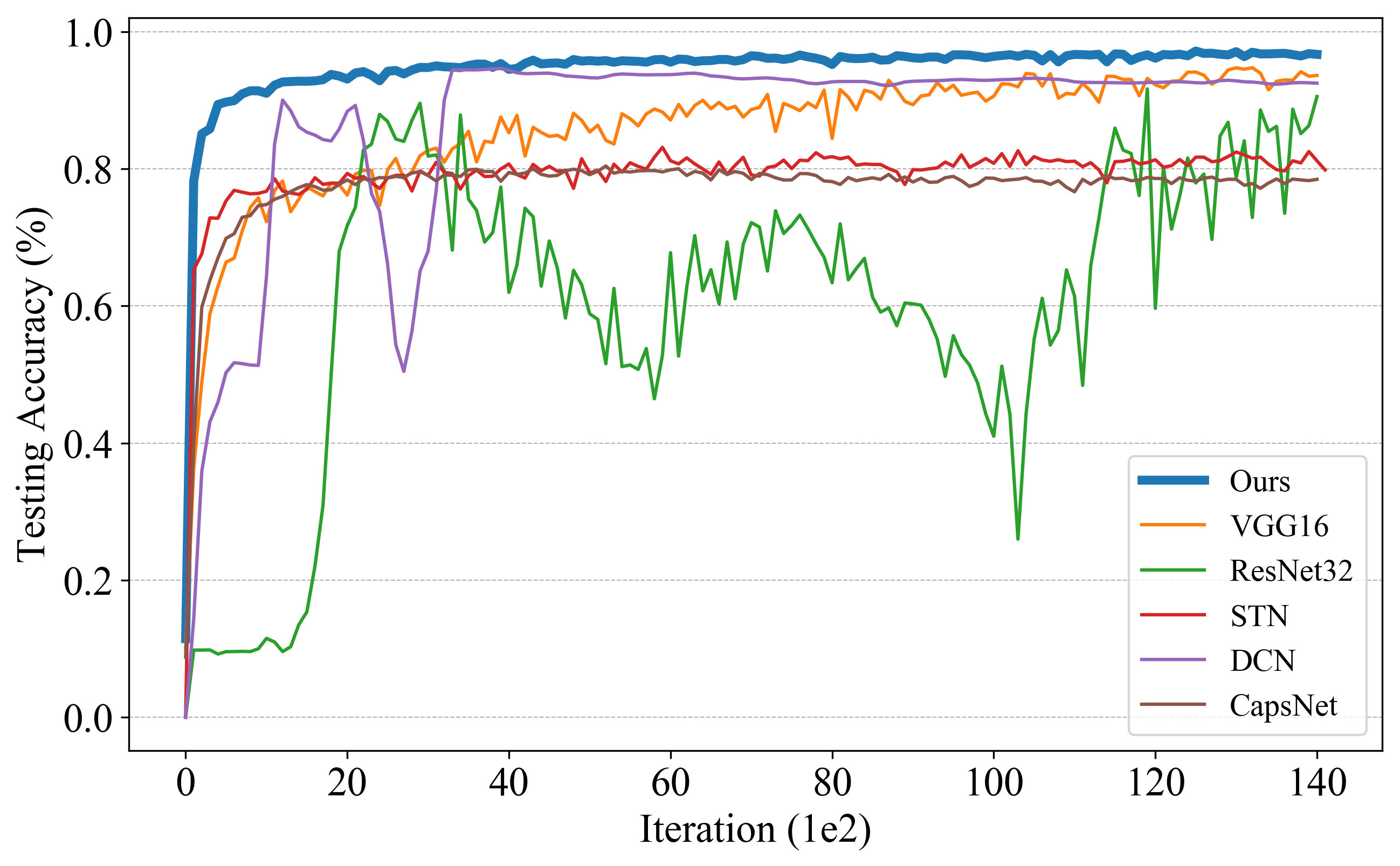}
     %\vspace{-10pt}     
   \end{minipage}   
   \begin{minipage}{0.495\textwidth}
     \centering
     \includegraphics[width=.85\linewidth]{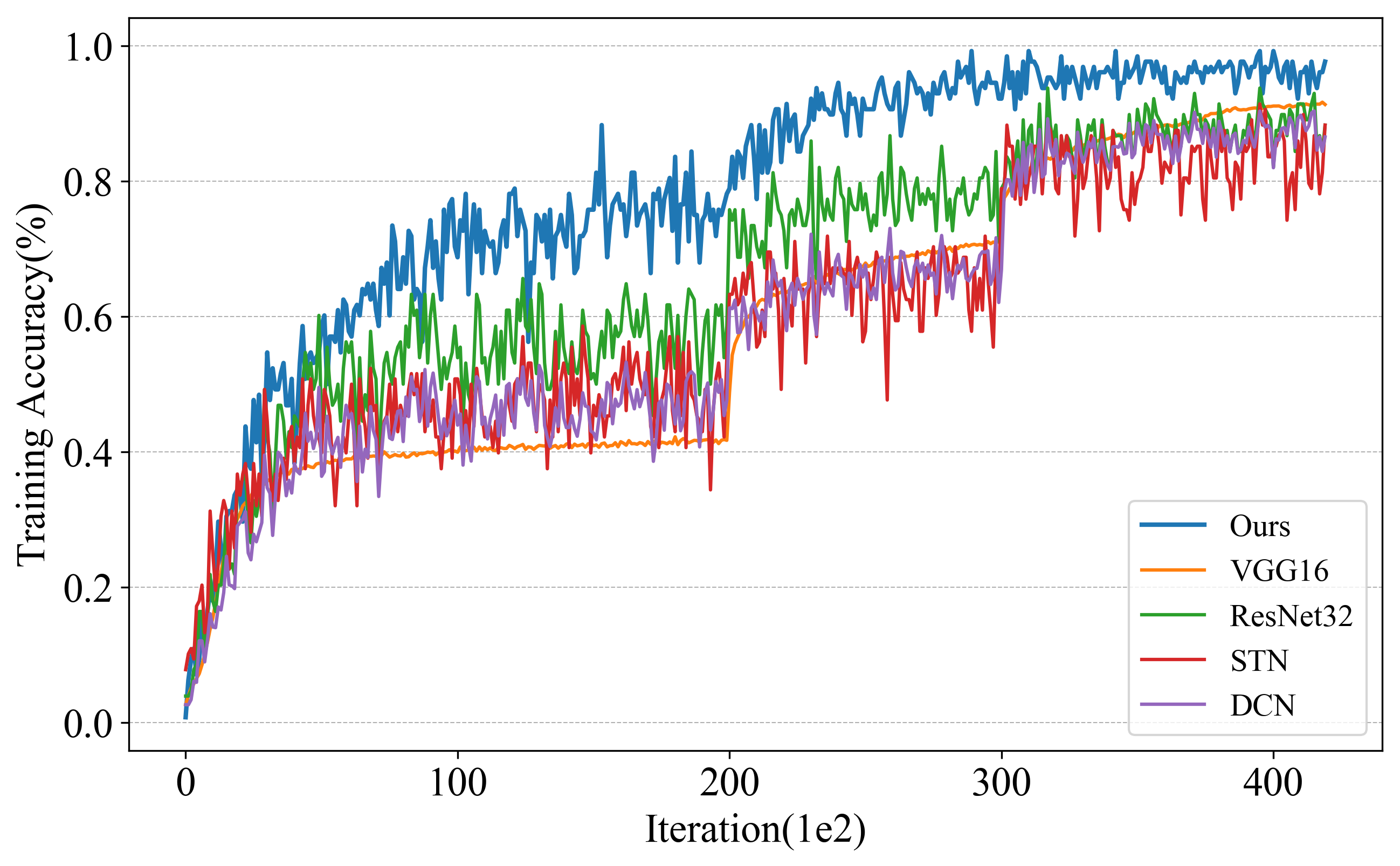}
     %\vspace{-10pt}     
   \end{minipage}\hfill
   \begin{minipage}{0.495\textwidth}
     \centering
     \includegraphics[width=.85\linewidth]{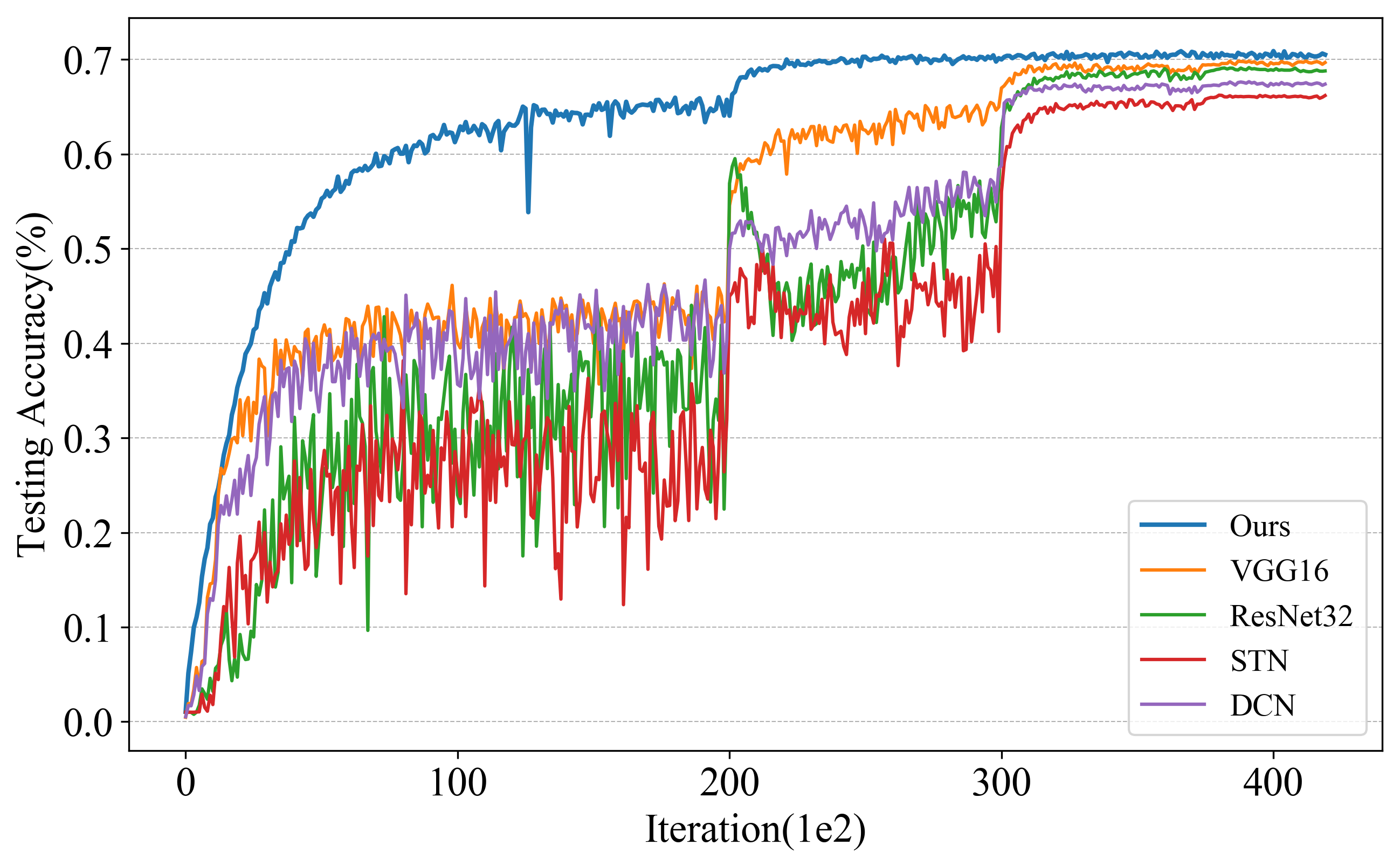}         
   \end{minipage} 
   \vspace{-1mm} 
   \caption{\footnotesize Illustration of training and testing accuracy on ({\bf top}) affNIST and ({\bf bottom}) CIFAR-100.}\label{fig:loss}
   \vspace{-3mm} 
\end{figure*}

\textbf{Training \& Testing Behavior:}
To validate our results, we show the training and testing accuracy behavior of each network on affNIST (full training dataset) and CIFAR-100, where each iteration contains 128 mini-batches. On affNIST we set learning rate to 0.001. On CIFAR-100, we start with learning rate $0.1$, divide it by 10 after 20K and 30K iterations, and terminate training after 42K iterations. As we see all the networks are well trained with convergence. In the testing stage our DPN converges much faster than the others, leading to big gaps in the first few iterations. Similar observations can be made in training as well. From this perspective, we can also demonstrate that DPN has better generalization than the other networks.

\setlength{\tabcolsep}{2pt}
\begin{wraptable}{r}{8cm}\small
	\vspace{-15pt}
	\begin{center}    
		\begin{tabular}{|c||c|c|c|c|c|c|}
			\hline & {\bf Ours} & VGG16 & ResNet32 & STN & DCN & CapsNet \\ 
            \hline affNIST & 37.8 & 14.7 & 47.3 & 12.7 & 83.0 & 65.0 \\
            \hline CIFAR-100 &  102 & 52.0 & 138.6 & 63.0 & 152.0 & - \\
			\hline
		\end{tabular}
	\end{center}
    \vspace{-2mm}
	\caption{\footnotesize Comparison on training time (s) per epoch.}\label{tab:time}
    \vspace{-3mm}
\end{wraptable}
We record the training time of each network from tensorboard and list them in Table \ref{tab:time}. Since the complexity of DM is the same as 2D convolution, our DPNs should be able to be trained as efficiently as other CNN based networks. Indeed we can verify this by comparing DPN17 with VGG16 and ResNet32. In addition DPN17 can be trained faster than DCN and CapsNet.

\section{Conclusion}
In this paper we propose novel Deformable Part Networks (DPNs) for 2D object recognition that can be interpreted as detecting objects within the networks for learning pose-invariant features. By comparing with some state-of-the-art networks, we demonstrate that empirically our DPNs can achieve better performance with better generalization and better tolerance to affine transformations, thanks to the learning of spatial configurations of deformable parts for pose estimation.

\newpage
{\small
\bibliographystyle{ieee}
\bibliography{egbib}

\begin{thebibliography}{10}\itemsep=-1pt

\bibitem{crandall2005spatial}
D.~Crandall, P.~Felzenszwalb, and D.~Huttenlocher.
\newblock Spatial priors for part-based recognition using statistical models.
\newblock In {\em CVPR}, volume~1, pages 10--17, 2005.

\bibitem{dai2017deformable}
J.~Dai, H.~Qi, Y.~Xiong, Y.~Li, G.~Zhang, H.~Hu, and Y.~Wei.
\newblock Deformable convolutional networks.
\newblock In {\em CVPR}, pages 764--773, 2017.

\bibitem{humanpose2015sp}
K.~Duan, D.~Batra, and D.~Crandall.
\newblock Human pose estimation through composite multi-layer models.
\newblock {\em Signal Processing}, 110:15--26, May 2015.

\bibitem{felzenszwalb2010cascade}
P.~F. Felzenszwalb, R.~B. Girshick, and D.~McAllester.
\newblock Cascade object detection with deformable part models.
\newblock In {\em CVPR}, pages 2241--2248, 2010.

\bibitem{felzenszwalb2010object}
P.~F. Felzenszwalb, R.~B. Girshick, D.~McAllester, and D.~Ramanan.
\newblock Object detection with discriminatively trained part-based models.
\newblock {\em TPAMI}, 32(9):1627--1645, 2010.

\bibitem{felzenszwalb2005pictorial}
P.~F. Felzenszwalb and D.~P. Huttenlocher.
\newblock Pictorial structures for object recognition.
\newblock {\em IJCV}, 61(1):55--79, 2005.

\bibitem{felzenszwalb2012distance}
P.~F. Felzenszwalb and D.~P. Huttenlocher.
\newblock Distance transforms of sampled functions.
\newblock {\em Theory Of Computing}, 8:415--428, 2012.

\bibitem{fischler1973representation}
M.~A. Fischler and R.~A. Elschlager.
\newblock The representation and matching of pictorial structures.
\newblock {\em IEEE Transactions on computers}, 100(1):67--92, 1973.

\bibitem{ghiasi2014occlusion}
G.~Ghiasi and C.~C. Fowlkes.
\newblock Occlusion coherence: Localizing occluded faces with a hierarchical
  deformable part model.
\newblock In {\em CVPR}, pages 2385--2392, 2014.

\bibitem{girshick2014rich}
R.~Girshick, J.~Donahue, T.~Darrell, and J.~Malik.
\newblock Rich feature hierarchies for accurate object detection and semantic
  segmentation.
\newblock In {\em CVPR}, pages 580--587, 2014.

\bibitem{girshick2015deformable}
R.~Girshick, F.~Iandola, T.~Darrell, and J.~Malik.
\newblock Deformable part models are convolutional neural networks.
\newblock In {\em CVPR}, pages 437--446, 2015.

\bibitem{goodfellow2013maxout}
I.~J. Goodfellow, D.~Warde-Farley, M.~Mirza, A.~Courville, and Y.~Bengio.
\newblock Maxout networks.
\newblock In {\em ICML}, pages III--1319, 2013.

\bibitem{he2015delving}
K.~He, X.~Zhang, S.~Ren, and J.~Sun.
\newblock Delving deep into rectifiers: Surpassing human-level performance on
  imagenet classification.
\newblock In {\em ICCV}, pages 1026--1034, 2015.

\bibitem{he2016deep}
K.~He, X.~Zhang, S.~Ren, and J.~Sun.
\newblock Deep residual learning for image recognition.
\newblock In {\em CVPR}, pages 770--778, 2016.

\bibitem{inverse-graphics}
G.~Hinton.
\newblock Taking inverse graphics seriously.
\newblock https://www.cs.toronto.edu/~hinton/csc2535/notes/lec6b.pdf.

\bibitem{hinton2018matrix}
G.~Hinton, N.~Frosst, and S.~Sabour.
\newblock Matrix capsules with em routing.
\newblock In {\em ICLR}, 2018.

\bibitem{hinton2011transforming}
G.~E. Hinton, A.~Krizhevsky, and S.~D. Wang.
\newblock Transforming auto-encoders.
\newblock In {\em International Conference on Artificial Neural Networks},
  pages 44--51. Springer, 2011.

\bibitem{ioffe2015batch}
S.~Ioffe and C.~Szegedy.
\newblock Batch normalization: Accelerating deep network training by reducing
  internal covariate shift.
\newblock In {\em ICML}, pages 448--456, 2015.

\bibitem{isik2013dynamics}
L.~Isik, E.~M. Meyers, J.~Z. Leibo, and T.~Poggio.
\newblock The dynamics of invariant object recognition in the human visual
  system.
\newblock {\em Journal of neurophysiology}, 111(1):91--102, 2013.

\bibitem{jaderberg2015spatial}
M.~Jaderberg, K.~Simonyan, A.~Zisserman, et~al.
\newblock Spatial transformer networks.
\newblock In {\em NIPS}, pages 2017--2025, 2015.

\bibitem{kingma2014adam}
D.~P. Kingma and J.~Ba.
\newblock Adam: A method for stochastic optimization.
\newblock {\em arXiv preprint arXiv:1412.6980}, 2014.

\bibitem{krizhevsky2009learning}
A.~Krizhevsky.
\newblock Learning multiple layers of features from tiny images.
\newblock 2009.

\bibitem{kumar2017pose}
V.~Kumar, A.~Namboodiri, M.~Paluri, and C.~Jawahar.
\newblock Pose-aware person recognition.
\newblock In {\em CVPR}, pages 6223--6232, 2017.

\bibitem{lecun1998mnist}
Y.~LeCun.
\newblock The mnist database of handwritten digits.
\newblock {\em http://yann. lecun. com/exdb/mnist/}.

\bibitem{maaten2008visualizing}
L.~v.~d. Maaten and G.~Hinton.
\newblock Visualizing data using t-sne.
\newblock {\em JMLR}, 9(Nov):2579--2605, 2008.

\bibitem{masi2016pose}
I.~Masi, S.~Rawls, G.~Medioni, and P.~Natarajan.
\newblock Pose-aware face recognition in the wild.
\newblock In {\em CVPR}, pages 4838--4846, 2016.

\bibitem{mhaskar2016deep}
H.~N. Mhaskar and T.~Poggio.
\newblock Deep vs. shallow networks: An approximation theory perspective.
\newblock {\em Analysis and Applications}, 14(06):829--848, 2016.

\bibitem{nair2010rectified}
V.~Nair and G.~E. Hinton.
\newblock Rectified linear units improve restricted boltzmann machines.
\newblock In {\em ICML}, pages 807--814, 2010.

\bibitem{ouyang2015deepid}
W.~Ouyang, X.~Wang, X.~Zeng, S.~Qiu, P.~Luo, Y.~Tian, H.~Li, S.~Yang, Z.~Wang,
  C.-C. Loy, et~al.
\newblock Deepid-net: Deformable deep convolutional neural networks for object
  detection.
\newblock In {\em CVPR}, pages 2403--2412, 2015.

\bibitem{sabour2017dynamic}
S.~Sabour, N.~Frosst, and G.~E. Hinton.
\newblock Dynamic routing between capsules.
\newblock In {\em NIPS}, pages 3859--3869, 2017.

\bibitem{simonyan2014very}
K.~Simonyan and A.~Zisserman.
\newblock Very deep convolutional networks for large-scale image recognition.
\newblock {\em arXiv preprint arXiv:1409.1556}, 2014.

\bibitem{szegedygoing}
C.~Szegedy, W.~Liu, Y.~Jia, P.~Sermanet, S.~Reed, D.~Anguelov, D.~Erhan,
  V.~Vanhoucke, A.~Rabinovich, J.-H. Rick~Chang, et~al.
\newblock Going deeper with convolutions.
\newblock In {\em CVPR}, 2015.

\bibitem{szegedy2016rethinking}
C.~Szegedy, V.~Vanhoucke, S.~Ioffe, J.~Shlens, and Z.~Wojna.
\newblock Rethinking the inception architecture for computer vision.
\newblock In {\em CVPR}, pages 2818--2826, 2016.

\bibitem{tian2012exploring}
Y.~Tian, C.~L. Zitnick, and S.~G. Narasimhan.
\newblock Exploring the spatial hierarchy of mixture models for human pose
  estimation.
\newblock In {\em ECCV}, pages 256--269. Springer, 2012.

\bibitem{torralba2004sharing}
A.~Torralba, K.~P. Murphy, and W.~T. Freeman.
\newblock Sharing features: efficient boosting procedures for multiclass object
  detection.
\newblock In {\em CVPR}, 2004.

\bibitem{wei2016convolutional}
S.-E. Wei, V.~Ramakrishna, T.~Kanade, and Y.~Sheikh.
\newblock Convolutional pose machines.
\newblock In {\em CVPR}, pages 4724--4732, 2016.

\bibitem{wu2016learning}
T.~Wu, B.~Li, and S.-C. Zhu.
\newblock Learning and-or model to represent context and occlusion for car
  detection and viewpoint estimation.
\newblock {\em PAMI}, 38(9):1829--1843, 2016.

\bibitem{yu2015multi}
F.~Yu and V.~Koltun.
\newblock Multi-scale context aggregation by dilated convolutions.
\newblock {\em arXiv preprint arXiv:1511.07122}, 2015.

\bibitem{zhang2016object}
Z.~Zhang and P.~H. Torr.
\newblock Object proposal generation using two-stage cascade svms.
\newblock {\em TPAMI}, 38(1):102--115, 2016.

\bibitem{zhu2010latent}
L.~Zhu, Y.~Chen, A.~Yuille, and W.~Freeman.
\newblock Latent hierarchical structural learning for object detection.
\newblock In {\em CVPR}, pages 1062--1069. IEEE, 2010.

\end{thebibliography}
}  
	
\end{document}